# AI for Cultural Heritage Textiles: Fine-Tuned Latent Diffusion for Novel Ulos Motif Synthesis

Humasak Tommy Argo Simanjuntak*

Information Systems Study Program, Faculty of Informatics and Electrical Engineering, Institut Teknologi Del, humasak@del.ac.id

Jesika Purba

Information Systems Study Program, Faculty of Informatics and Electrical Engineering, Institut Teknologi Del, iss21049@students.del.ac.id

Sitogab Girsang

Information Systems Study Program, Faculty of Informatics and Electrical Engineering, Institut Teknologi Del, iss21015@students.del.ac.id

Widya Manurung

Information Systems Study Program, Faculty of Informatics and Electrical Engineering, Institut Teknologi Del, iss21039@students.del.ac.id

Samuel Situmeang

Information Systems Study Program, Faculty of Informatics and Electrical Engineering, Institut Teknologi Del, samuel.situmeang@del.ac.id

Arlinta Barus

Informatics Study Program, Faculty of Informatics and Electrical, Institut Teknologi Del, arlinta@del.ac.id

Daniel Oranova Siahaan

Informatics Department, Institut Teknologi Sepuluh Nopember, daniel@if.its.ac.id

Preserving and revitalising traditional textiles such as Ulos, a cultural heritage of the Batak ethnic group in North Sumatra, Indonesia, requires balancing fidelity to tradition with innovative approaches that meet contemporary design demands. Traditional Ulos weaving faces two key limitations: a narrow range of motifs and a time-intensive design process. This study presents a generative AI framework that fine-tunes two pretrained latent diffusion models: Protogen v3.4 and Stable Diffusion v1.4, on a curated, annotated dataset of high-resolution Ulos motifs to generate culturally consistent yet novel designs. Model

* Place the footnote text for the author (if applicable) here.

performance is evaluated quantitatively using Fréchet Inception Distance (FID), Inception Score (IS), and qualitatively through assessments by traditional weavers and members of the public. Protogen v3.4 consistently outperforms Stable Diffusion v1.4, achieving substantially lower FID (~10.5×) and higher IS (2.0×), indicating superior visual fidelity, diversity, and closer alignment with the real Ulos motif distribution. We further examine the effects of strength and guidance scale on generation quality across both models. Lower strength values consistently yield higher fidelity (lower FID), while higher strength values increase generative diversity at the cost of realism, revealing a clear fidelity–diversity trade-off for both models. Across all tested configurations, a guidance scale of 5–9 provides the most effective balance between fidelity and diversity, stabilising FID, KID, and IS, and is recommended as the operating range for high-quality, diverse Ulos motif generation. These findings demonstrate that carefully fine-tuned generative AI can support the creative renewal of intangible cultural heritage while preserving its stylistic and symbolic integrity.

## 1 INTRODUCTION

Weaving, a timeless art form, has been an integral part of human civilisation, serving as a means of cultural expression and a tool for preserving identity. Traditional weaving practices worldwide have produced intricate patterns and motifs that encode their respective communities' stories, beliefs, and values. With its rich diversity of motifs and techniques, Indonesia's traditional weaving is a source of global recognition and pride. Over fifty distinct textile traditions, each originating from a unique region from Sumatra to Papua, such as Ulos, Songket, Ikat, Tenun, and many more, are deeply enmeshed in cultural identity and artisanal heritage [1]. These weaves, not just artistic expressions, but also represent intangible cultural heritage, are a testament to the human capacity for creativity and skill. The fact that UNESCO has recognised several Indonesian weaving traditions as significant contributions to global cultural diversity is a testament to their value and a source of pride for Indonesians.

Ulos weaving, a vital cultural tradition for the Batak ethnic group in North Sumatra, plays a crucial role in preserving their unique cultural identity. More than just a woven cloth, Ulos is imbued with deep symbolic meaning, functioning in ceremonial, social, and spiritual contexts, and representing a cultural heritage passed down through generations [2, 3]. Each Ulos motif carries a distinct philosophical message and visual characteristics, but the traditional weaving requires considerable skill, time, and adherence to established design conventions. For example, Figure 1a shows Ulos Harungguan, a Toba Batak ethnic ulos weave known as the "king of ulos" because it combines all other ulos motifs in one cloth, symbolising harmony and unity. This cloth holds profound significance in traditional ceremonies, is often worn by leaders, and symbolises prayers for their blessing with wisdom and prosperity. Ulos Harungguan is intricately crafted and takes considerable time to weave, but it is now a symbol of Batak cultural pride, respected and cherished. Another picture, Figure 1 (b), shows a traditional Batak weaver weaving Ulos using a "Gedogan". This simple device places one end of the warp on a pole and the other on the weaver's belt, allowing the weaver to control the yarn's tension. This practice emphasises the continuity of ancestral weaving techniques, which are still maintained in the production of Ulos with distinctive motifs.

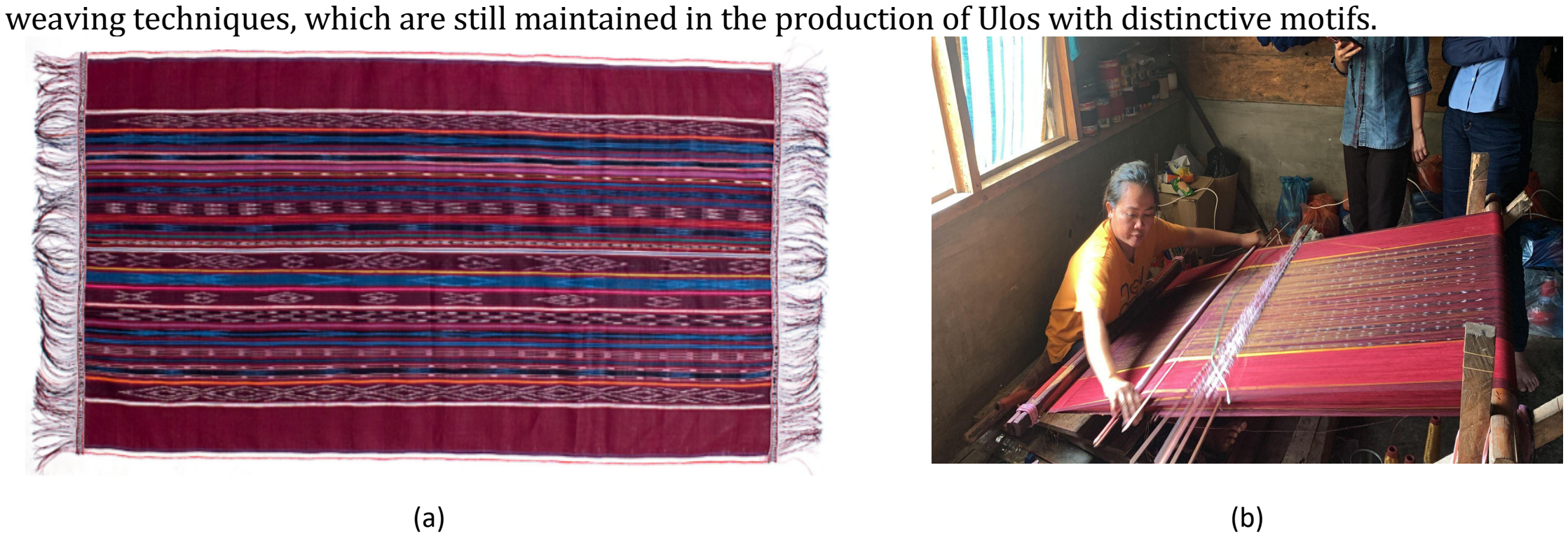

(a) (b)

Figure 1: The example of ulos harungguan (a), traditional weaving with Gedogan (b)

Preserving Ulos weaving while encouraging design innovation is a complex challenge in the era of globalisation. Preserving the authenticity of traditional motifs is key to maintaining cultural integrity; on the other hand, introducing new designs can revitalise the local weaving industry, expand market share, and attract the younger generation. This challenge is further exacerbated by the limited variety of motifs in traditional Ulos weaving and the high time and skill investment required to design and produce new motifs. These conditions complicate the innovation process for responding to evolving market trends, as any changes must remain aligned with the craft's philosophical values and cultural identity.

The development of Generative Artificial Intelligence (AI) in recent years has opened significant new opportunities to address the challenges of preserving and innovating motif designs in traditional weaving. This technology enables the creation of new visual content by learning from existing data, allowing AI to automatically mimic, modify, and combine elements of complex visual styles. Generative models, particularly image-based models such as Generative Adversarial Networks (GANs) and latent diffusion models, can learn the in-depth characteristics of a visual style from curated datasets, including geometric shapes, colour palettes, and pattern structures, and then generate new designs that maintain aesthetic coherence with the source while introducing creative variations [4, 5, 6, 7, 8, 9]. In the context of cultural heritage preservation, this capability not only enables the accurate digital reproduction of traditional motifs but also ensures that the symbolic value and cultural philosophy they embody are respected, providing reassurance about the preservation of our cultural heritage.

Previous studies in automatic Ulos motif generation have primarily utilised the DiTenun platform, which applies two core generative approaches: image quilting and GAN. The image quilting method, while effective for generating digital motifs by synthesising larger textures from the rearrangement of smaller patches from original images [10], has significant limitations when applied to complex cultural motifs. The method often produces visible repetitions, with patterns that recur in fixed directions, reducing the naturalness of the generated design. In addition, imperfect patch selection frequently results in unnatural seams or abrupt transitions, disrupting visual coherence and lowering the overall quality of the output. More critically, image quilting lacks the ability to generalise beyond simple, stochastic textures such as sand, grass, or basic fabric grain. When confronted with motifs that contain significant structural and symbolic elements, such as Ulos, the method fails to preserve semantic consistency and cultural integrity. Its structural inflexibility prevents the creation of new geometric forms or innovative arrangements, while artefacts and redundancy become more apparent at larger scales. Consequently, although image quilting provides a basic framework for motif synthesis, it is insufficient for generating novel, complex, and culturally meaningful weaving patterns.

Complementing the image quilting method, DiTenun has implemented SinGAN, a GAN model that learns from a single input image to generate new motifs with similar visual features [11]. SinGAN's strength lies in its ability to generate new, coherent images that retain the style and patterns of the original. By focusing on local statistics (patches), SinGAN captures textures, shapes, and structures of the single training image. SinGAN learns the pattern of one image through multi-scale GAN training, then uses this knowledge to generate endless realistic variations. Although SinGAN has demonstrated the ability to create motif variations, its output quality and diversity remain heavily dependent on the richness of the input. When the input lacks complexity or variation, the resulting motifs often display poor diversity and visual artefacts, particularly at larger scales or with intricate structures. Furthermore, SinGAN's stochasticity impedes control over specific motif characteristics during generation. It's excellent for textures and patterns but struggles with large, complex object structures, such as the ulos motif.

DiTenun's latest research turned to StyleGAN, which supports multiple input images and employs mixing regularisation to blend their features, yielding more diverse, more controllable, and more realistic image synthesis, as well as high-resolution motif outputs [5, 12, 13, 14]. The model merges shapes, patterns, and colours, generating motifs that incorporate elements from multiple Ulos while still producing new designs. The StyleGAN produced new Ulos motifs with altered patterns (shifts in motif direction, missing or additional shapes), discolouration (blended colours from multiple input motifs), and merged motifs (two or more motifs combined into a single new design) [15]. While StyleGAN can mix and merge motifs, sometimes uncontrolled alterations of the output result in a loss of symbolic structure (the traditional balance of lines, symmetry, or geometry is not preserved), discolour motifs in culturally inaccurate ways, and produce artefacts when combining complex motifs. This is problematic because Ulos designs often carry deep cultural meaning, and even small motif distortions can alter their symbolic value. Furthermore, StyleGAN is computationally intensive, especially during training for Ulos motifs, which require long iterations (100k–1.2M) and substantial GPU resources, which may be unrealistic for small cultural research groups.

The main limitation of previous approaches, which we must urgently address, lies in their inability to produce truly varied Ulos motifs that remain culturally authentic. While methods such as image quilting, SinGAN, and StyleGAN have demonstrated the capacity to synthesise new motifs, their effectiveness is constrained by structural artefacts, repetitive patterns, and limited semantic control. StyleGAN, in particular, achieves impressive visual fidelity [15], yet its reliance on large, diverse training datasets, coupled with the difficulty of preserving the symbolic structures inherent to Ulos motifs, often results in outputs that lack cultural coherence. These shortcomings highlight the need for a more adaptive generative paradigm that ensures visual quality and diversity and aligns more closely with the symbolic and aesthetic rules embedded in traditional design.

Recent advances in diffusion-based generative models offer a compelling alternative to overcome these challenges. The Latent Diffusion Model (LDM), as implemented in the widely adopted Stable Diffusion framework [6] [16, 17, 18, 19, 20, 21], performs the generative process in a compressed latent space, enabling both computational efficiency and high-resolution synthesis. As a specific, open-source implementation of LDM, Stable Diffusion conditioned on text embeddings via the CLIP model has gained significant traction for text-to-image generation across creative and scientific domains. More importantly, LDM provides stronger semantic control through prompt-based or conditional generation, allowing for richer variations of motifs while preserving cultural authenticity. This enhanced controllability and adaptability make diffusion models particularly suited for Ulos motif generation, enabling the creation of novel designs that respect traditional structures while expanding creative possibilities. In this work, we explore the application of LDM to Ulos motifs, demonstrating its potential to advance digital cultural preservation, a crucial aspect of our work, and innovation beyond the capabilities of previous GAN-based approaches.

In summary, this study advances the state of generative modelling for cultural heritage through three principal contributions. First, we introduce a fine-tuned Latent Diffusion Model (LDM) tailored to Ulos motifs, enabling the generative process to capture structural fidelity while embedding the symbolic semantics intrinsic to these designs. Second, we demonstrate the capacity of generative AI to synthesise entirely new motifs that extend beyond existing repositories, offering both creative diversity and cultural coherence. Third, we situate this work within the broader agenda of digital cultural preservation, illustrating how diffusion-based generative models can serve as instruments of visual innovation and mechanisms for safeguarding, revitalising, and transmitting intangible heritage. Collectively, these contributions establish diffusion-driven approaches as a promising paradigm for reconciling computational creativity with cultural continuity in the digital age.

## 2 METHOD

The methodological workflow comprises five main stages: (i) Data Collection: curating 88 traditional Ulos motif fragments from various sources; (ii) Data Preparation: resizing all images to 512×512 pixels, and augmenting the dataset to 299 samples to ensure balanced ethnic representation and varied visual complexity; (iii) Model Training: adopts a generative image synthesis method based on the Latent Diffusion Model (LDM) framework [6], specifically employing and fine-tuning two pretrained latent diffusion models (Stable Diffusion v1.4 and Protogen x3.4) on the prepared dataset; (iv) Model Testing: generating images across scenarios of structural variation, color variation, and high-complexity input generation; and (vi) Evaluation: assessing outputs through quantitative metrics (Fréchet Inception Distance, Inception Score), and qualitative validation from weavers and the public participant.

### 2.1 Research Dataset

The primary material for this research is a curated dataset of Ulos motif images, originally compiled by DiTenun (Digital Tenun Nusantara), an AI-driven textile innovation platform that creates digital woven motifs by combining technology and cultural traditions, dedicated to the digital preservation of Indonesian woven heritage. This dataset, referenced in prior academic studies [15], consists of digitised fragments of traditional Batak Ulos motifs obtained through archival documentation and direct collaboration with local weavers. Building on this foundation, the present study further enriched the dataset by incorporating additional materials from open-access cultural repositories and field visits to weaving communities across North Sumatra.

Rather than employing complete Ulos textiles, this study focuses on isolated motif fragments, representing the most minor repeatable design units, consistent with the motif-based classification framework previously established in Ulos textile research. The dataset comprises traditional Ulos motif fragments collected from three Batak sub-ethnic groups: Toba, Karo, and Simalungun, sourced via the DiTenun platform. Each motif is systematically categorised according to origin, ulos type, naming code, and the number of available motif variations. The compiled datasets are summarised in Table 1.

Table 1: Batak Ulos Motif Dataset

| Origin | Ulos Type | Motif Code | #Motif |
|---|---|---|---|
| Toba | Suri–suri | SS | 1 |
| Toba | Sadum Angkola | SA | 11 |
| Toba | Sumbat | S | 3 |
| Toba | Pinucca | PNC | 23 |
| Toba | Mangiring simareurreur | MS | 3 |
| Toba | Marpisoran | MP | 4 |
| Toba | Bintang Maratur | BM | 3 |
| Toba | Bolean | BL | 7 |
| Toba | Sibolang | BLG | 1 |
| Karo | Uis | U | 6 |
| Karo | Julu | JL | 3 |
| Karo | Bekan Buluh | BB | 2 |
| Karo | Sigara-gara | SGR | 2 |
| Karo | Gundur | GDR | 1 |
| Karo | Indung Bayu | IB | 2 |
| Karo | Ragi Barat | RB | 4 |
| Karo | Uis Teba | TD | 4 |

Figure 2 presents the curated Ulos motif dataset, comprising 88 unique digital images sourced from three Batak sub-ethnic groups: 56 Toba Batak ulos motifs, 24 Karo Batak ulos motifs, and 8 Simalungun Batak motifs. To compensate for the relatively small dataset size and to increase the variety and quantity of data, a comprehensive data augmentation pipeline is implemented, inspired by augmentation practices in low-resource generative modelling [22].

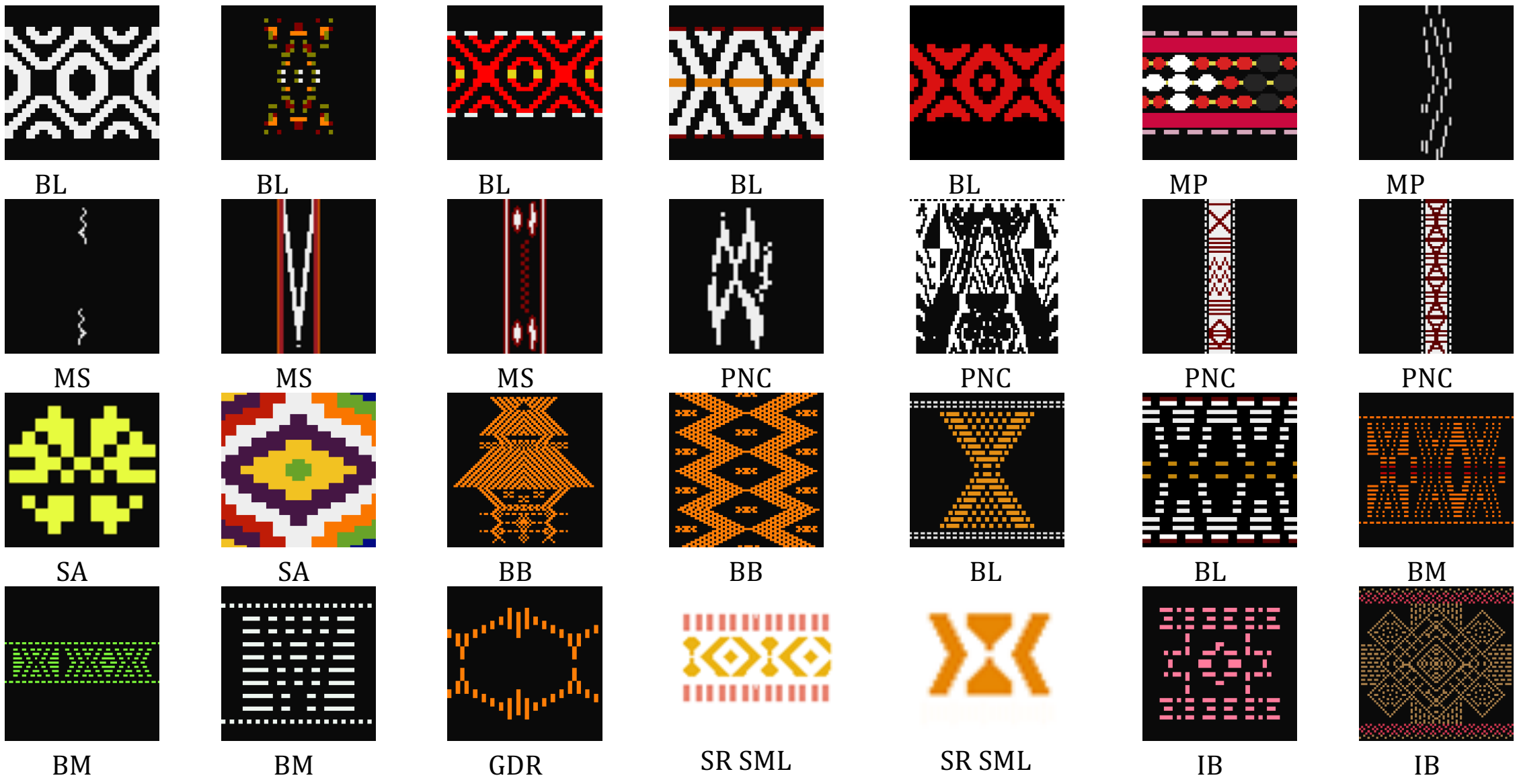

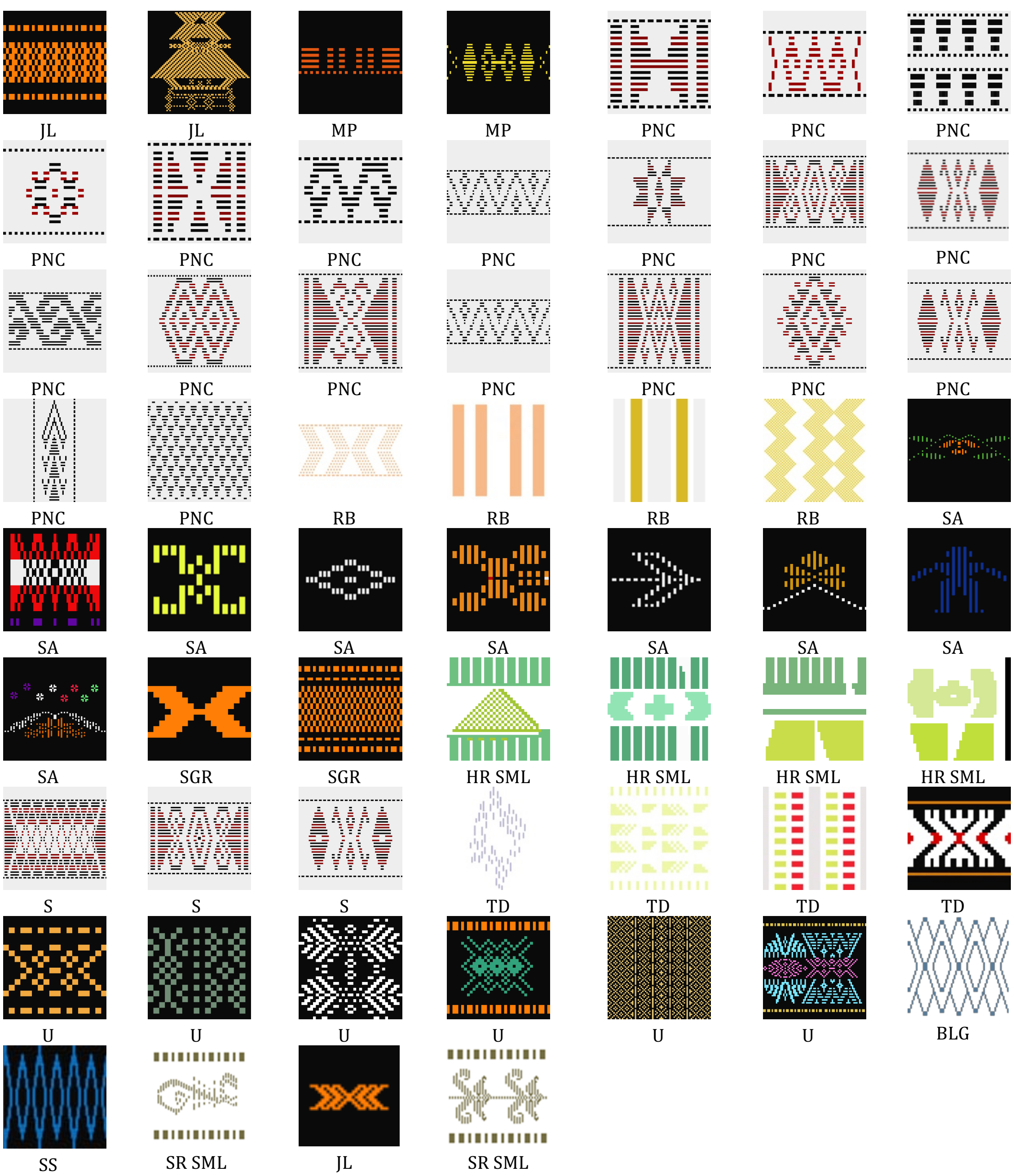

Figure 2: Training Dataset of Ulos Motif Images for Model Fine-Tuning

## 2.2 Methodological Workflow

In the data collection phase, a curated dataset of 88 Ulos motif fragments is assembled from multiple sources, including the DiTenun project and open-access cultural archives. These fragments reflected the diversity of Batak sub-ethnic groups: Toba, Simalungun, and Karo, and included ceremonial motifs and contemporary reinterpretations. This selection ensured balanced cultural representation and visual variation, essential for robust generative modelling.

During data preparation, all images are standardized to 512×512 pixels to meet the input requirements of the latent diffusion architecture. To increase dataset diversity without compromising structural authenticity, augmentation techniques such as ±30° random rotation, horizontal and vertical flipping, and ±15% brightness variation are applied. A targeted classification scheme is also introduced to organize the dataset by cultural and visual complexity. Each motif is manually reviewed and categorised by density (sparse, moderate, dense), symmetry type (axial, radial, asymmetric), and border intricacy (presence or absence of ceremonial edge frames). This structured curation provided the model with a semantically rich and structurally varied training set.

For model training, two pre-trained latent diffusion models, Stable Diffusion v1.4 and Protogen x3.4 are fine-tuned on the curated dataset using an image-to-image strategy. Fine-tuning is guided by conditioning, a descriptive textual prompt designed to reflect the symbolic elements of Ulos design (e.g., "traditional Ulos motif with dense symmetrical weave and ceremonial borders"). Training is conducted for 50 epochs with a batch size of 2, a learning rate of $1.0\times10^{-5}$, steps 1000, size 512, and the Adam optimiser. Prompt conditioning enabled the models to reinterpret input motifs in line with semantic cues while retaining core cultural features. In the testing phase, model performance is evaluated under controlled experimental conditions that vary conditioning strength (0.65, 0.75, 0.85), guidance scale (5-19), and input complexity (ranging from simple motifs to dense and intricate compositions). This experimental setup enables a systematic assessment of the models' generative capabilities in terms of flexibility, realism, and cultural coherence. A summary of the testing configuration and dataset used for Stable Diffusion v1.4 (SD-v1.4) and Protogen x3.4 is provided in Table 2.

Table 2: Testing Scenario of Latent Diffusion Model

| #S | #I | Dataset | | Size | Learning Rate | Strength | | Guidance Scale | |
|---|---|---|---|---|---|---|---|---|---|
| | | Ulos Type | Category | | | Min | Max | Min | Max |
| 1 | 1 | Toba | Bintang Maratur | 512 | 1.0e-5 | 0.65 | 0.85 | 5 | 19 |
| 1 | 2 | Karo | Uis | 512 | 1.0e-5 | 0.65 | 0.85 | 5 | 19 |
| 1 | 3 | Simalungun | Hati Rongga | 512 | 1.0e-5 | 0.65 | 0.85 | 5 | 19 |
| 2 | 1 | Toba | Bintang Maratur | 512 | 1.0e-5 | 0.65 | 0.85 | 5 | 19 |
| 2 | 2 | Karo | Uis | 512 | 1.0e-5 | 0.65 | 0.85 | 5 | 19 |
| 2 | 3 | Simalungun | Hati Rongga | 512 | 1.0e-5 | 0.65 | 0.85 | 5 | 19 |
| 3 | 1 | Karo | Uis | 512 | 1.0e-5 | 0.65 | 0.85 | 5 | 19 |
| 3 | 2 | Karo | Uis | 512 | 1.0e-5 | 0.65 | 0.85 | 5 | 19 |

Each testing scenario targets a distinct aspect of generative control: Scenario 1 evaluates structural manipulation, assessing the model's ability to alter motif shape and arrangement while preserving cultural characteristics; Scenario 2 evaluates colour variation, where only the colour composition changes while structure and layout remain intact; and Scenario 3 evaluates complex motif handling, testing robustness in generating dense, detailed patterns without losing clarity or authenticity.

**Scenario 1: Shape and Motif Structure Modification**

This scenario explores the model's capacity to reconfigure the structural layout of traditional motifs. The objective is to assess whether the model can propose new visual arrangements, such as changes in motif alignment, spacing, and orientation, while preserving the cultural essence embedded in the geometric forms of Ulos textiles. The model is conditioned using the following prompts:

**Positive Prompt:**

"*A high-resolution Batak Ulos textile with symmetrical vertical layout, featuring woven star, square, rectangle, and floral motifs arranged in a pixel-perfect structure. Each motif is sharp and detailed, with thread textures and traditional vertical borders, blending geometric elegance with cultural structure*."

**Negative Prompt:**

"*green, orange, yellow, abstract symbols, blurry, low resolution, fuzzy, painterly, glitch, messy, tribal shapes, cartoon, hard outlines, glowing, chaotic patterns, JPEG artefacts, deformed shapes*."

**Scenario 2: Colour Variation**

The second experimental scenario investigates the model's capability to perform precise chromatic transformations while preserving the original geometric integrity and spatial composition of Batak Ulos motifs. Unlike the previous scenario, this experiment constrains the generative process exclusively to colour modification, enforcing strict adherence to the source motif's structure, proportions, and alignment. The objective is to assess the Latent Diffusion Model's (LDM) ability to produce culturally faithful colour adaptations, particularly within the traditional palette of deep reds, blacks, and off-whites, without introducing structural distortion or semantic drift. The model is conditioned through controlled text prompts designed to isolate colour transformation as the only modifiable attribute:

**Positive Prompt:**

"*A high-resolution Batak Ulos textile that strictly preserves the exact pixel-based motif structure, geometric vertical layout, and solid black or white background from the original image. The only modification applied is the recolouring into deep, culturally rich shades of red, such as crimson and dark maroon. The woven motifs, including stars, squares, and linear patterns, must remain completely unchanged in shape, scale, alignment, and position.*"

**Negative Prompt:**

"*shape change, motif distortion, structure change, layout shift, background colour change, motif replacement, new motifs, added elements, blurry, low resolution, glowing, distortion, glitch, painterly, cartoon, tribal abstraction, pattern drift, colour bleed, broken symmetry, jagged edges, style transfer*."

**Scenario 3: High Complexity Motif**

The third scenario examines the model's robustness to visually dense, high-complexity inputs. These motifs typically exhibit layered symmetry, multi-colour compositions, and intricate geometric forms, resembling entirely woven textile segments rather than isolated motif fragments. This scenario serves as a stress test for the generative model, assessing its ability to handle dense visual information while maintaining motif integrity and aesthetic clarity.

**Positive Prompt:**

"*A high-resolution Batak Ulos textile with symmetrical vertical layout, featuring woven star, square, rectangle, and floral motifs arranged in a pixel-perfect structure. Each motif is sharp and detailed, with thread textures and traditional vertical borders, blending geometric elegance with cultural structure*."

**Negative Prompt:**

"*green, orange, yellow, abstract symbols, blurry, low resolution, fuzzy, painterly, glitch, messy, tribal shapes, cartoon, hard outlines, glowing, chaotic patterns, JPEG artefacts, deformed shapes.*"

Finally, a multi-perspective evaluation is conducted, combining quantitative and qualitative assessments. Quantitatively, Fréchet Inception Distance (FID) and Inception Score (IS) are used to evaluate image realism, diversity, and distributional alignment with the original dataset. Qualitatively, three experienced Ulos weavers are interviewed to assess the feasibility of weaving, cultural fidelity, and motif integrity. In addition, an online survey with 30 general participants is conducted to evaluate aesthetic appeal and cultural resonance. This dual evaluation ensured rigorous validation of both technical performance and cultural authenticity.

### 2.3 Latent Diffusion Model Pipeline

The generative framework employed in this study is the Latent Diffusion Model (LDM), proposed by Rombach et al. [6] and developed by CompVis (https://github.com/CompVis/latent-diffusion). Unlike pixel-based diffusion models, which are computationally expensive due to their reliance on high-dimensional image space, LDM performs denoising in a compressed latent space learned from a pretrained Variational Autoencoder (VAE). This separation between perceptual compression and generative modelling allows LDM to retain semantic detail while significantly reducing computational costs.

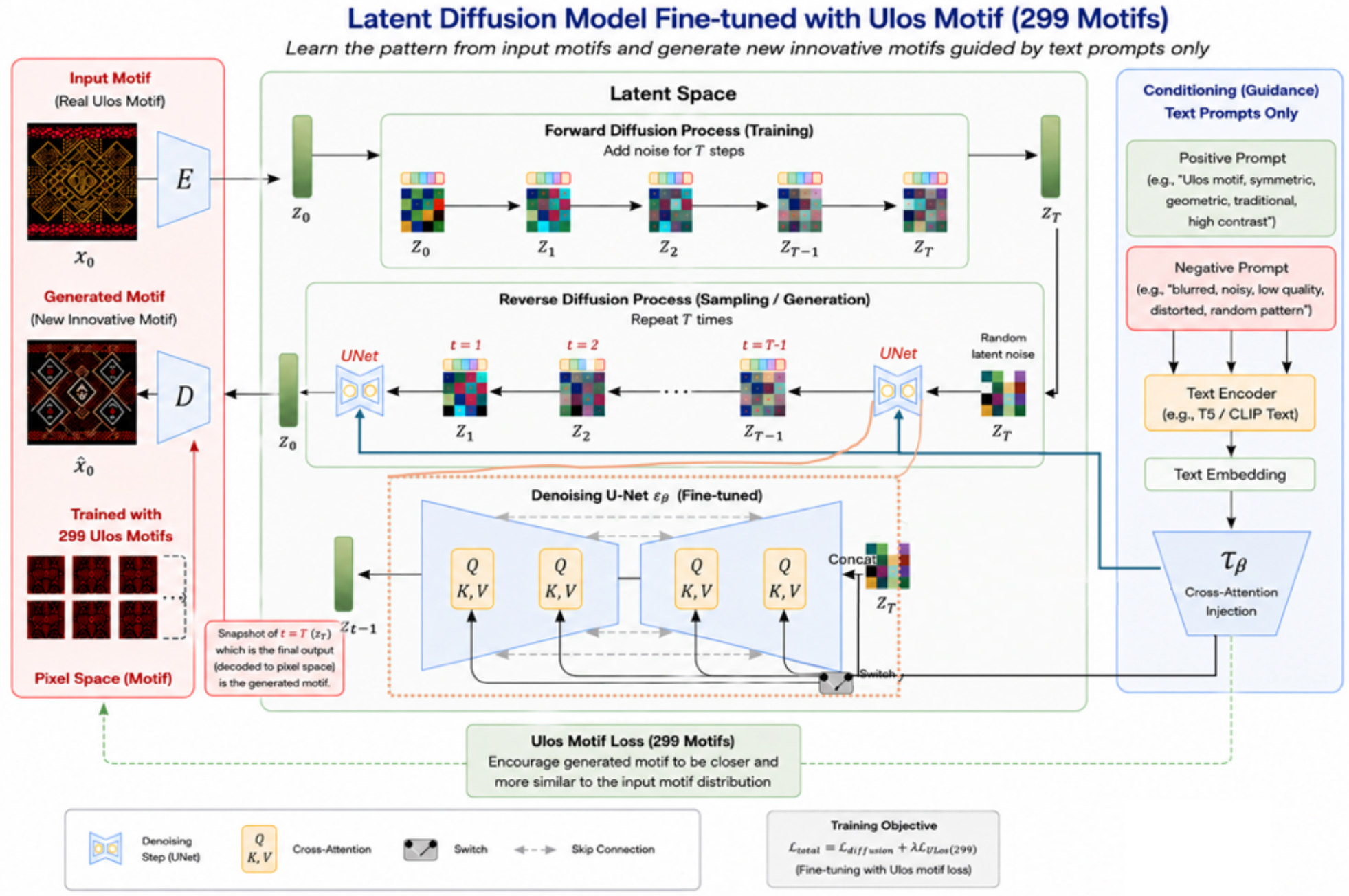

Figure 3: Architecture of the fine-tuned Latent Diffusion Model (LDM) for novel Ulos motif synthesis, illustrating the forward diffusion process, text-conditioned reverse diffusion via cross-attention, and the Ulos motif loss used during fine-tuning on 299 motifs.

As illustrated in Figure 3, the architecture consists of four principal components: (i) Encoding, where ulos motif images are mapped into a lower-dimensional latent representation; (ii) Diffusion, where a sequence of denoising steps progressively refines noisy latent variables toward coherent representations; (iii) Conditioning, implemented through cross-attention mechanisms that integrate auxiliary information such as text features to guide generation; and (iv) Decoding, where the refined latent vectors are reconstructed into high-resolution images in pixel space.

This design achieves a balance between efficiency, controllability, and fidelity, enabling high-resolution image synthesis and multimodal conditioning beyond the capabilities of earlier generative models such as GANs.

The equation (1) presented below extends the Latent Diffusion Model (LDM) framework introduced by Rombach et al. [6]. Specifically, they build on the conditional LDM objective (Equation 2–3 of the original paper) and the cross-attention-based conditioning mechanism, adapting both to fine-tune a text-conditioned LDM on a curated dataset of 299 Batak Ulos motifs, as illustrated in the accompanying architecture diagram (Figure 3).

$$L_{total}(\theta, \tau_\theta) = E_{\mathcal{E}(x), y, \varepsilon \sim N(0,1), t}[\|\varepsilon - \varepsilon_\theta(z_t, t, \tau_\theta(y))\|_2^2] + \lambda \cdot E_{\hat{z}_0 \sim p\theta}\left[min_{i\,=\,1,\ldots,299}\, d\left(D(\hat{z}_0), x^{(i)}\right)\right] \quad (1)$$

Equation (1) extends the baseline objective by introducing an additional Ulos motif loss term, λ·L_Ulos(299), corresponding to the "Ulos Motif Loss (299 Motifs)" component shown in Figure 3. The first term, L_diffusion, retains the original noise-prediction objective of Rombach et al. unchanged, while the second term regularizes generation by penalizing the perceptual or feature-space distance d(·,·) between the decoded estimate of the model's predicted clean latent, $D(\hat{z}_0)$, and its nearest neighbour among the 299 real Ulos motifs $x^{(i)}$ used for fine-tuning. As defined in Equation. (2), d(·,·) is a perceptual feature distance, consistent with the perceptual loss that Rombach et al. themselves use to train their autoencoder; this choice captures structural and stylistic similarity between the generated and authentic Ulos motifs more robustly than a raw pixel-space distance such as L2 or L1. The weighting coefficient λ balances generative novelty against fidelity to authentic Ulos structure, allowing the fine-tuned model to synthesize new motifs that remain culturally and geometrically coherent rather than drifting toward the unconstrained, general-purpose outputs of the original LDM.

$$d(u, v) := \|\varphi(u) - \varphi(v)\|_2^2 \quad (2)$$

Equation (2) defines d(·,·): φ(·) is a feature map from a fixed, pretrained network, and d(u,v) is the squared Euclidean distance between the feature representations of two images. This follows the perceptual loss used by Rombach et al. A feature-space distance allows the Ulos motif loss to tolerate minor pixel-level variation while still penalizing motifs that depart structurally from authentic Ulos patterns.

$$\tilde{\varepsilon}_\theta(z_t, t, y) = \varepsilon_\theta(z_t, t, \emptyset) + s \cdot (\varepsilon_\theta(z_t, t, \tau_\theta(y)) - \varepsilon_\theta(z_t, t, \emptyset)) \quad (3)$$

Equation (2) extends the cross-attention conditioning mechanism to the sampling stage by incorporating classifier-free guidance, the mechanism underlying the positive and negative prompt branches in the "Conditioning (Guidance)" module of Figure 3. Here, $\varepsilon\theta(z_t, t, \emptyset)$ denotes the unconditional noise prediction obtained from the empty or negative prompt (e.g., "blurred, noisy, low quality, distorted, random pattern"), while $\varepsilon\theta(z_t, t, \tau\theta(y))$ is the prediction conditioned on the positive textual prompt y (e.g., "Ulos motif, symmetric, geometric, traditional, high contrast"). The guidance scale s linearly amplifies the difference between these two predictions, which has the effect of maximizing the alignment of the generated image with the intended text description: larger values of s push the sampled motif more strongly toward the prompt-described structure and colour characteristics, at the cost of reduced sample diversity, consistent with the guidance-scale sweep (s = 5 to 19) and the corresponding FID instability observed in the scenario-based evaluations.

Overall, the Latent Diffusion Model (LDM) for generating Ulos motifs can be understood as a two-stage process: a visual transformation and a semantic creative process. In the first stage, the traditional Ulos motif undergoes a significant transformation from pixel space to latent space. This transformation is facilitated by the encoder-decoder, a key component that represents the visual meaning of the Ulos motif in a more compact form. This form still preserves the motif structure, fabric texture, and distinctive Batak symmetrical patterns. In the latent space, the model performs a diffusion process, gradually adding and removing noise while learning to recognise weaving patterns, geometric shapes, and distinctive colours (red, black, and white). This process allows for the exploration of modern design variations without losing the inherent cultural identity of Ulos.

The second stage is the creative process, driven by a denoising U-Net and a conditioning system. The U-Net removes noise while preserving the details of the motif structure. However, it is the cross-attention mechanism that plays a key role in this process. This mechanism connects the visual form with semantic context, such as text ("modern symmetric ulos with star patterns") or reference images of traditional motifs. By doing so, it ensures that the generative results respect symbolic meanings, such as courage, loyalty, and brotherhood, while allowing for modern touches in the design's colour, rhythm, and proportions. This emphasis on the cross-attention mechanism underscores its role in preserving the cultural significance of the Ulos motifs, making LDM a powerful tool that combines technology and cultural values.

A central contribution of this study is integrating a conditioning mechanism within the diffusion framework to align generative outputs with cultural and visual semantics. During training, paired text–image data are employed, where each Ulos motif image is annotated with descriptive captions that convey structural, stylistic, and symbolic attributes (e.g., motif layout, weaving structure, or cultural meaning). These captions are transformed into dense vector representations using a frozen transformer-based text encoder, typically derived from the CLIP model. The resulting embeddings are injected into the denoising U-Net via cross-attention layers, in which text embeddings serve as Keys (K) and Values (V), while image feature maps act as Queries (Q). This architecture facilitated multi-scale alignment between semantic prompts and visual features, thereby guiding the reverse diffusion process to produce motif variations faithful to their intended cultural and aesthetic descriptions [23] [24].

In inference, conditioning is performed using manually authored prompts that specify stylistic preferences, motif characteristics, or cultural constraints. In addition to positive prompts, negative prompts are introduced to suppress undesired properties such as blurring, asymmetry, or artefacts incompatible with traditional Ulos weaving. For example, a representative prompt is:

"A high-resolution Batak Ulos textile with a symmetrical vertical layout, featuring woven star, square, rectangle, and floral motifs arranged in a pixel-perfect structure. Each motif is sharp and detailed, with thread textures and traditional vertical borders, blending geometric precision with cultural symbolism."

Such conditioning enhanced visual fidelity and reinforced cultural authenticity, ensuring that generative outputs remained aesthetically compelling and contextually appropriate [25]

## 3 RESULTS AND DISCUSSION

This section presents the quantitative and qualitative evaluation results of the text-conditioned Latent Diffusion Models for Ulos motif generation across three testing scenarios. The results are analysed using FID and IS metrics, complemented by participant and traditional weaver assessments to evaluate visual quality, cultural authenticity, and practical weaving feasibility.

### 3.1 Training Result

The results of the training phase are presented in this section, highlighting the performance of the fine-tuned latent diffusion models on the curated Ulos dataset. The analysis focuses on how Stable Diffusion v1.4 and Protogen x3.4 adapted their generative capabilities to capture structural fidelity, cultural symbolism, and stylistic variation within traditional motifs. Comparative results are reported to evaluate convergence behaviour, visual quality, and the models' ability to preserve the distinctive characteristics of Batak textile design. Figure 4 illustrates the training loss curves for Protogen x3.4 and Stable Diffusion v1.4 (SD-v1.4) over 50 epochs on the curated Ulos motif dataset

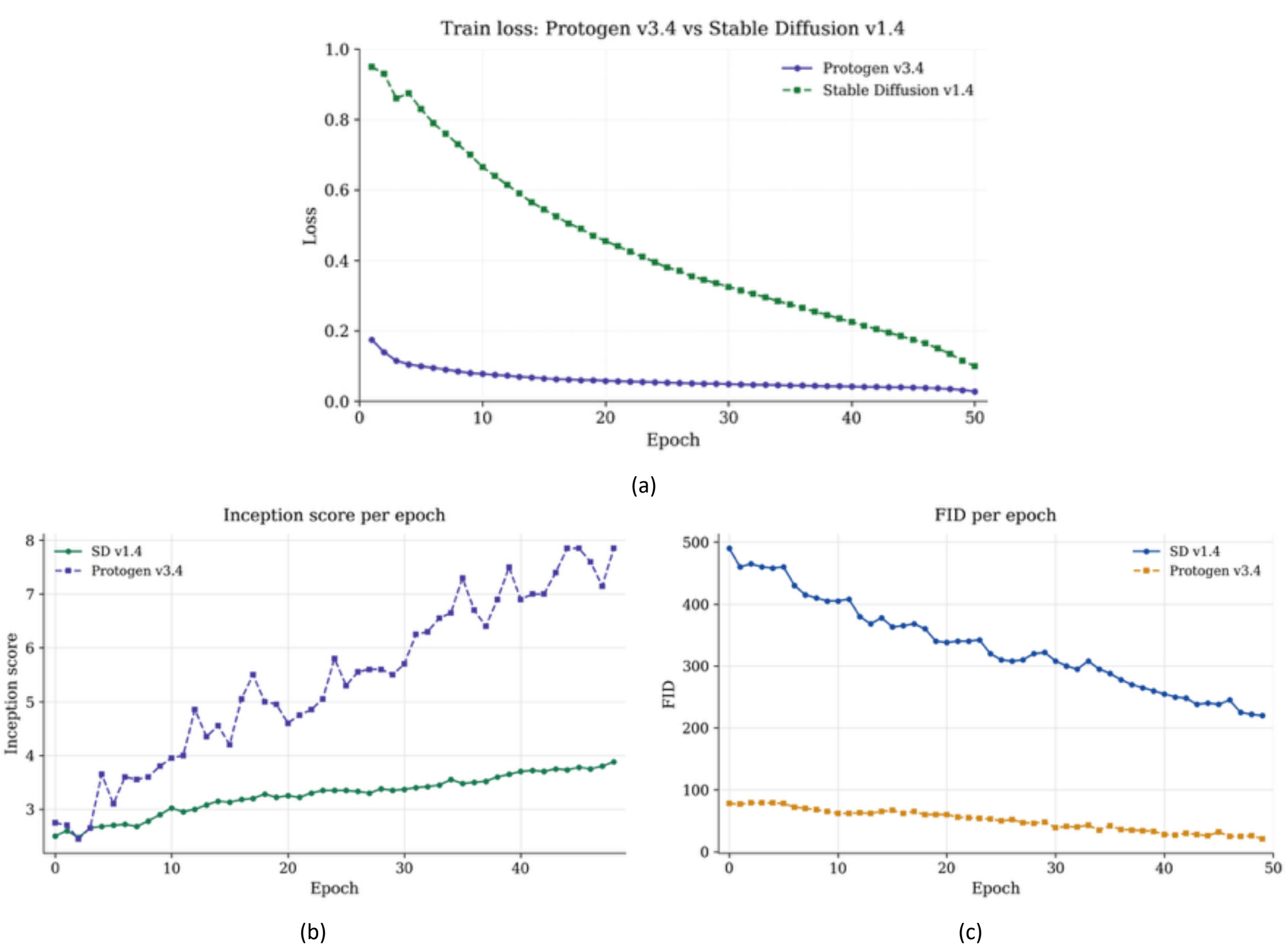

Figure 4: Training performance of Protogen v3.4 and Stable Diffusion v1.4 during Ulos motif fine-tuning: (a) training loss, (b) Inception Score, and (c) Fréchet Inception Distance across 50 epochs.

The fine-tuning trajectories of the two backbone models reveal markedly different convergence behaviours when adapted to the Ulos motif dataset. Protogen v3.4 exhibited rapid loss minimization, declining from an initial value of approximately 0.18 to below 0.05 within the first ten epochs and stabilizing near 0.03 by epoch 50, indicating efficient adaptation to the target domain with minimal residual reconstruction error. In contrast, Stable Diffusion v1.4 began training from a substantially higher loss (~0.95) and exhibited a slower, near-exponential decay, plateauing at approximately 0.10 after 50 epochs, nearly three times Protogen v3.4's loss. This disparity in convergence is reflected in the Inception Score (IS), where Protogen v3.4 showed a markedly steeper upward trajectory, rising from ~2.7 to a peak of ~7.8–7.9, compared to Stable Diffusion v1.4's more modest, gradual improvement from ~2.5 to ~3.9. The substantially higher and more volatile IS trend observed for Protogen v3.4 suggests that its fine-tuned outputs possess greater feature diversity and higher per-image confidence in classifier predictions, likely attributable to Protogen's pretrained prior on stylized and illustrative imagery, which provides a more favourable initialization for capturing the geometric and decorative complexity inherent to Ulos motifs.

The Fréchet Inception Distance (FID) further corroborates the superiority of Protogen v3.4 in generating Ulos motifs that closely approximate the real data distribution. Stable Diffusion v1.4 exhibited a high initial FID of ~490, decreasing only to ~220 by the final epoch, indicating a persistent and substantial divergence between the feature statistics of generated and authentic Ulos motifs even after extended fine-tuning. Protogen v3.4, by comparison, began at a considerably lower FID (~78) and converged to ~21, representing roughly a tenfold improvement in distributional alignment relative to Stable Diffusion v1.4's final value. Taken together, the consistently lower loss, higher Inception Score, and substantially lower FID achieved by Protogen v3.4 indicate that it not only learns the target domain more efficiently but also produces motifs that are both visually diverse and statistically closer to authentic Ulos patterns. These findings suggest that Protogen v3.4 constitutes a more suitable backbone for fine-tuning latent diffusion models on culturally specific, structurally intricate textile motifs, likely due to its richer stylistic prior, whereas Stable Diffusion v1.4's general-purpose pretraining appears less optimized for capturing the fine-grained symbolic and geometric regularities characteristic of Ulos weaving patterns.

### 3.2 Testing Result

The testing phase is designed to evaluate the generative performance of the fine-tuned diffusion models beyond training metrics, focusing on their ability to synthesize culturally coherent, visually diverse Ulos motifs under varying experimental conditions. Table 3 and Table 4 present representative samples from the curated Ulos dataset used for testing, along with the corresponding outputs generated by the latent diffusion model. Each image pair illustrates how the model interprets and reconstructs the visual semantics of Ulos motifs under varying design conditions. The samples are systematically categorized by motif complexity and scale, enabling a structured comparative evaluation across the three experimental scenarios.

Table 3: Sample Ulos motif generated by testing the latent diffusion model for each scenario 1 and 2

| Input | 1 | | 2 | |
|---|---|---|---|---|
| | SDV 1.4 | Protogen x3.4 | SDV 1.4 | Protogen x3.4 |
| | | | | |
| | | | | |

Table 4: Sample Ulos motif generated by testing the latent diffusion model for scenario 3

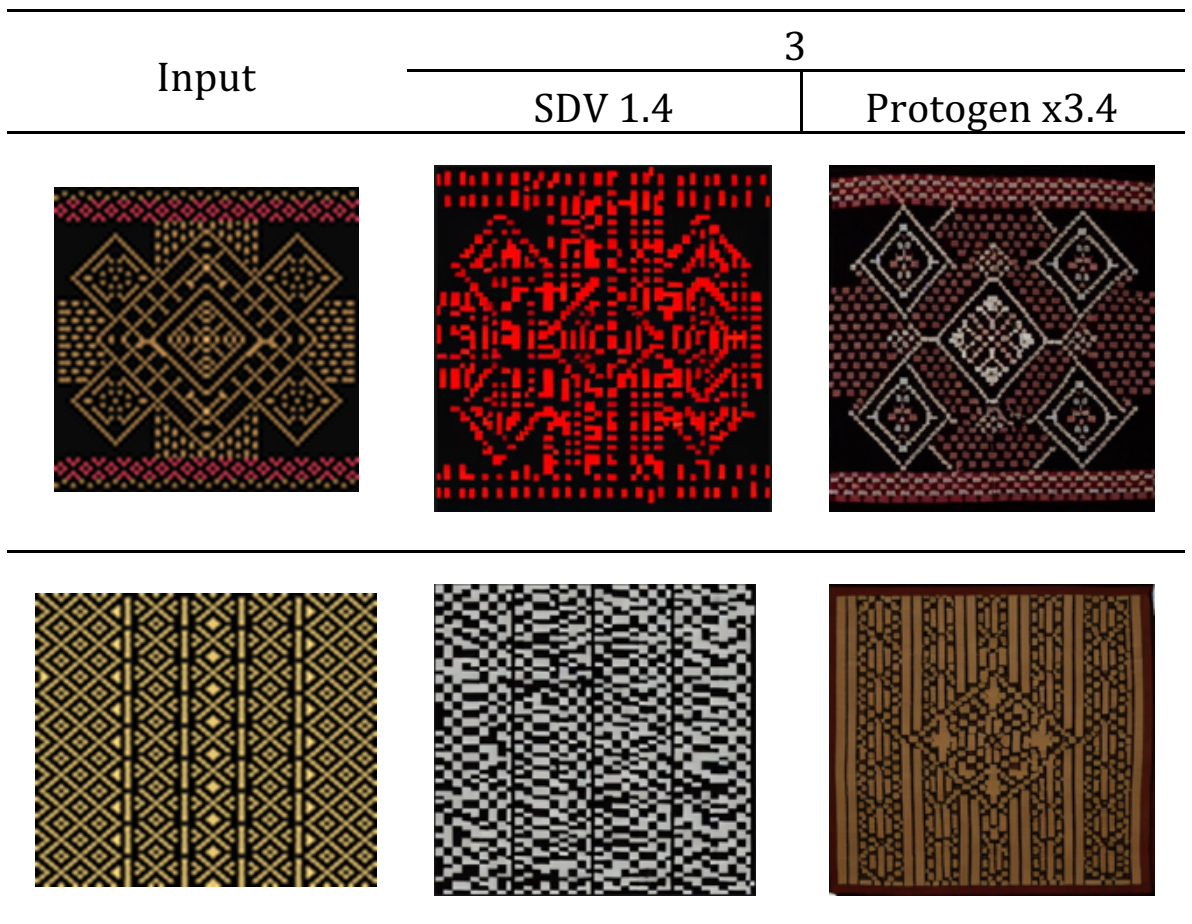

### *3.2.1 Quantitative Evaluation.*

Following the testing phase, each experimental scenario is quantitatively assessed to measure the generative performance of the latent diffusion models. Each model is evaluated across three strength settings (0.65, 0.75, and 0.85) and eight guidance scale values ranging from 5 to 19. The strength parameter controls the degree of deviation from the original input image, thereby influencing the level of transformation introduced during generation. In contrast, the guidance scale regulates the model's adherence to the semantic and stylistic constraints encoded within the conditioning prompt.

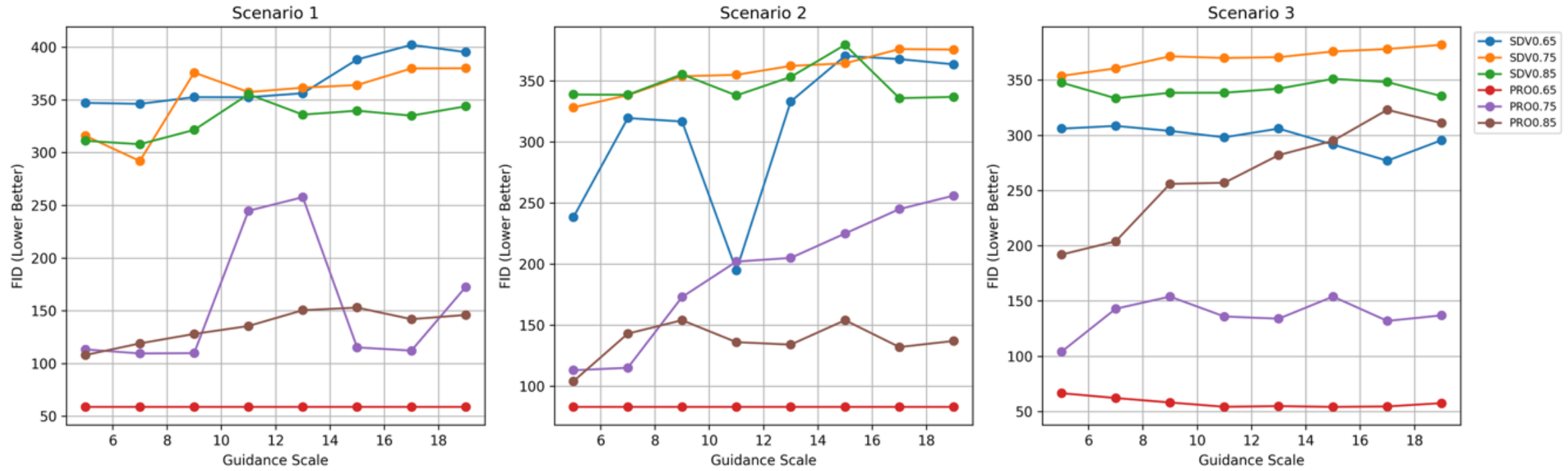

Figure 5: FID vs. guidance scale comparison between Protogen v3.4 and Stable Diffusion v1.4 across testing scenarios

Figure 5 shows the Fréchet Inception Distance (FID) as a function of guidance scale for both Protogen v3.4 (PRO) and Stable Diffusion v1.4 (SDV) across three evaluation configurations, strength settings of 0.65, 0.75, and 0.85, applied to each of the three testing scenarios: structural manipulation (Scenario 1), colour variation (Scenario 2), and complex motif handling (Scenario 3). The FID results show a clear performance gap between SDV 1.4 and Protogen x3.4. Across the three scenarios, Protogen x3.4 generally produces FID values much lower than SDV 1.4. This indicates that the feature distribution of the generated Ulos motifs from Protogen x3.4 is closer to that of the original Ulos motif dataset. The consistently lower FID values obtained by Protogen x3.4 suggest that the generated motifs have visual structures, textures, and geometric patterns that are closer to the original Ulos motifs. This means the generated motifs are not only visually plausible but also statistically aligned with the real motif dataset.

For Scenario 1, Protogen x3.4 with strength 0.65 achieves a stable FID value of 58.8 across all guidance scales. In contrast, SDV 1.4 produced much higher FID values, mostly above 300. This suggests that Protogen x3.4 learned the Ulos motif distribution more effectively and is less sensitive to changes in guidance scale. For Scenario 2, Protogen x3.4 again shows better FID performance than SDV 1.4, especially at strength 0.65. SDV 1.4 shows unstable behaviour, where the FID fluctuated strongly depending on the guidance scale. This instability indicates that SDV 1.4 may have difficulty maintaining consistent motif structure when the text-guidance strength changes. And for Scenario 3, Protogen x3.4 with strength 0.65 achieved the best overall FID values, ranging from 54 to 66. This is the strongest evidence that Protogen x3.4 can generate Ulos motifs that are visually and statistically close to the real data. Meanwhile, SDV 1.4 remained relatively high, indicating a larger mathematical distance between generated and real motif distributions. Overall, the FID analysis shows that Protogen x3.4 produces generated motifs that are much closer to the original Ulos motif distribution than SDV 1.4.

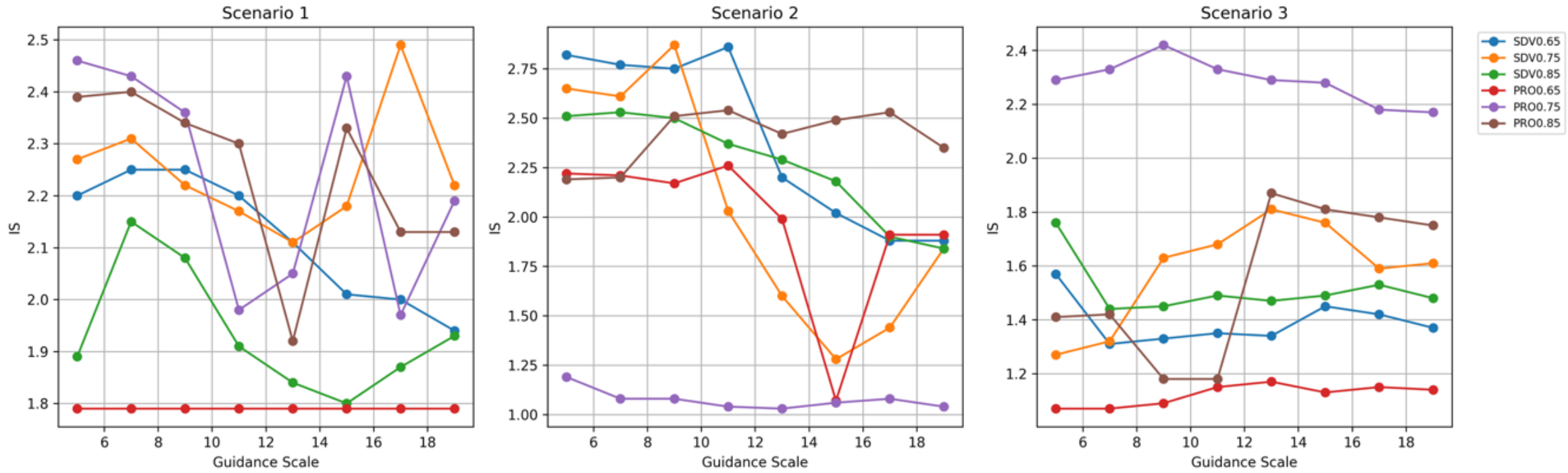

Figure 6: IS vs. guidance scale comparison between Protogen v3.4 and Stable Diffusion v1.4 across testing scenarios

Figure 6 shows the Inception Distance (IS) as a function of guidance scale for both Protogen v3.4 (PRO) and Stable Diffusion v1.4 (SDV) across three evaluation configurations. The Inception Score evaluates two important aspects of generated images: recognizability and diversity. A higher IS usually indicates that the generated images are clearer, more classifiable, and more diverse.

In Scenario 1, which focuses on shape and motif-structure modification, both models produced moderate IS values. SDV 1.4 showed relatively stable IS values, especially at strengths 0.65 and 0.75, with several values above 2.0. This suggests that the generated motifs are generally recognizable and contain visible structural elements. Protogen x3.4 achieved its strongest IS in this scenario at strength 0.75 and 0.85, indicating that the generated motifs are visually clear and diverse when moderate to higher strength values are applied. However, Protogen x3.4 at strength 0.65 produces a constant IS of 1.79, suggesting more stable but less diverse outputs.

In Scenario 2, which focuses on colour variation, SDV 1.4 achieved some of the highest IS values, especially under strengths 0.65 and 0.75. This indicates that the model generated images with strong visual separability and recognizable pattern differences. However, these higher IS values should be interpreted with caution, as IS does not assess whether the generated motifs remain close to the authentic Ulos distribution. A high IS may indicate that the outputs are visually distinct, but not necessarily culturally faithful or structurally accurate. Protogen x3.4 showed mixed behaviour in this scenario. At a strength of 0.85, it achieves relatively strong IS values, suggesting better visual clarity and diversity. In contrast, at strength 0.75, IS values are low, indicating reduced recognizability or limited variation in the generated image set.

In Scenario 3, which uses high-complexity motifs, the IS results show a different pattern. Protogen x3.4 at strength 0.75 achieved relatively high IS values, especially compared with strength 0.65. This suggests that increasing the strength parameter can yield more diverse, visually distinct outputs when the input motif is complex. However, higher diversity may also increase the risk of deviating from the authentic motif structure. SDV 1.4 produced lower IS values in this scenario than in Scenario 2, suggesting it struggled more with dense, complex motif structures. This indicates that complex Ulos patterns may be more difficult for SDV 1.4 to generate clearly and consistently.

Overall, the IS results suggest that both SDV 1.4 and Protogen x3.4 are capable of producing visually recognizable Ulos-like motifs, but their behaviour differs across scenarios and strength values. SDV 1.4 sometimes achieves higher IS values, especially in Scenario 2, suggesting stronger visual diversity or more easily classifiable outputs. However, this does not necessarily mean that SDV 1.4 generated more authentic Ulos motifs. Protogen x3.4, while sometimes producing lower IS values, demonstrated more controlled behaviour across several settings and should be interpreted alongside FID and KID.

The combined analysis of FID and IS demonstrates that Protogen x3.4 consistently outperformed Stable Diffusion v1.4 (SDV 1.4) for Ulos motif generation. Although SDV 1.4 occasionally achieved higher IS values, indicating greater visual diversity and recognizability, its FID scores remained substantially higher across all scenarios. This suggests that while SDV 1.4 can generate visually varied motifs, the generated images are less similar to the authentic Ulos motif distribution. In contrast, Protogen x3.4 consistently produces much lower FID values, indicating a smaller mathematical distance between the feature distributions of real and generated motifs. At the same time, Protogen x3.4 maintains competitive IS values, demonstrating that the generated motifs remain visually recognizable while preserving the structural and cultural characteristics of traditional Ulos textiles. This balance between fidelity and diversity is particularly important for the generation of cultural heritage, where maintaining motif authenticity is more critical than maximizing visual variation alone.

Among all experimental settings, Protogen x3.4 achieves the best overall performance in Scenario 3, with a strength value of 0.65 and a guidance scale of 15, yielding the lowest FID score of 54.2 while maintaining an acceptable IS value of 1.13. Although some SDV 1.4 configurations produce higher IS scores, none achieved comparable FID performance. Therefore, considering both distributional similarity (FID) and image quality/diversity (IS), Protogen x3.4 provides the most effective and reliable model for generating novel Ulos motifs that are both visually coherent and culturally faithful to the original motif distribution.

*3.2.2 Qualitative Evaluation.*

The qualitative evaluation assesses the perceptual quality and cultural appropriateness of the generated Ulos motifs using two complementary approaches. First, in-depth interviews with traditional weavers, to capture expert insights regarding motif authenticity, aesthetic coherence, alignment with customary design principles, and the possibility of weaving the motif. Second, a public questionnaire is distributed to gather broader perceptions of the motifs' visual appeal, uniqueness, and overall acceptance from a non-expert audience. The combination of these methods ensures a comprehensive understanding, reinforcing the evaluation's thoroughness and reliability for the audience.

- Evaluation of Weaver’s Perception of the motif generation result

To assess the cultural and technical appropriateness of the generated motifs, a qualitative evaluation is conducted through structured interviews with three experienced traditional Ulos weavers. Each weaver is presented with a set of images produced by two generative models: Stable Diffusion v1.4 and Protogen x3.4, across three transformation configurations (strength: 0.65, 0.75, and 0.85; guidance scale: 5–19). They are asked to evaluate each motif along three dimensions: structural composition, clarity of the woven elements, and feasibility of realisation using traditional weaving techniques.

Following the interviews, all three weavers consistently selected specific generated images from each scenario that, in their view, still accurately represented Ulos design conventions and remained technically feasible to weave using manual methods. The most favoured images across all scenarios are displayed in Figure 7.

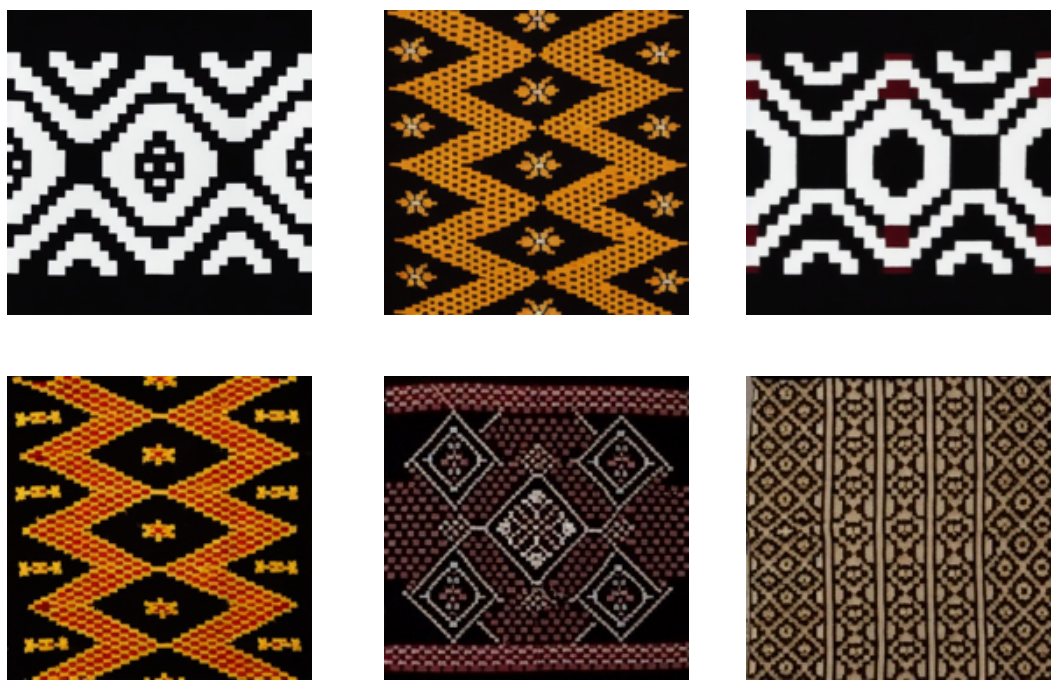

Figure 7: Top-rated LDM-generated motif images chosen by the weavers.

Structural elements such as horizontal alignment, visual symmetry, and proportional repetition are core characteristics in traditional Ulos motifs. According to the weavers, the motifs generated by Protogen x3.4 at moderate strength settings (0.65–0.75) demonstrated visual balance, clean motif segmentation, and thread paths that make the motif technically feasible to weave. Conversely, motifs generated by Stable Diffusion v1.4, particularly at a strength of 0.85, drew considerable criticism. Several images are deemed structurally confusing due to broken symmetry or vertical alignment that contradicts the customary horizontal flow of Ulos weaving. One weaver remarked that such vertical patterns would require additional "jumlah lidi" (warp spacers), significantly increasing complexity in loom setup and thread counting. The motifs are characterised as 'technically unweavable' and misaligned with traditional weaving orientation, suggesting a fundamental departure from conventional weaving principles.

In summary, this evaluation underscores that Protogen x3.4, under well-controlled configuration, better respects the visual semantics and material constraints of traditional Ulos weaving. It is considered more culturally sensitive, technically grounded, and functionally realistic. On the other hand, Stable Diffusion v1.4 requires additional prompt tuning and stricter visual conditioning to produce motifs that align with artisan expectations. This highlights the necessity of incorporating culturally informed human feedback, rather than just algorithmic performance, into the validation of generative models for heritage-driven design applications.

- Evaluation of Public Responses

To evaluate and compare the quality of images generated by the Ulos motif generation system based on Stable Diffusion, a questionnaire is distributed to 30 local participants with varying levels of familiarity with Ulos motifs.

The results of this evaluation are used to assess how well the system can generate new motif images that are symmetrical, attractive, and retain the traditional characteristics of Ulos motifs.

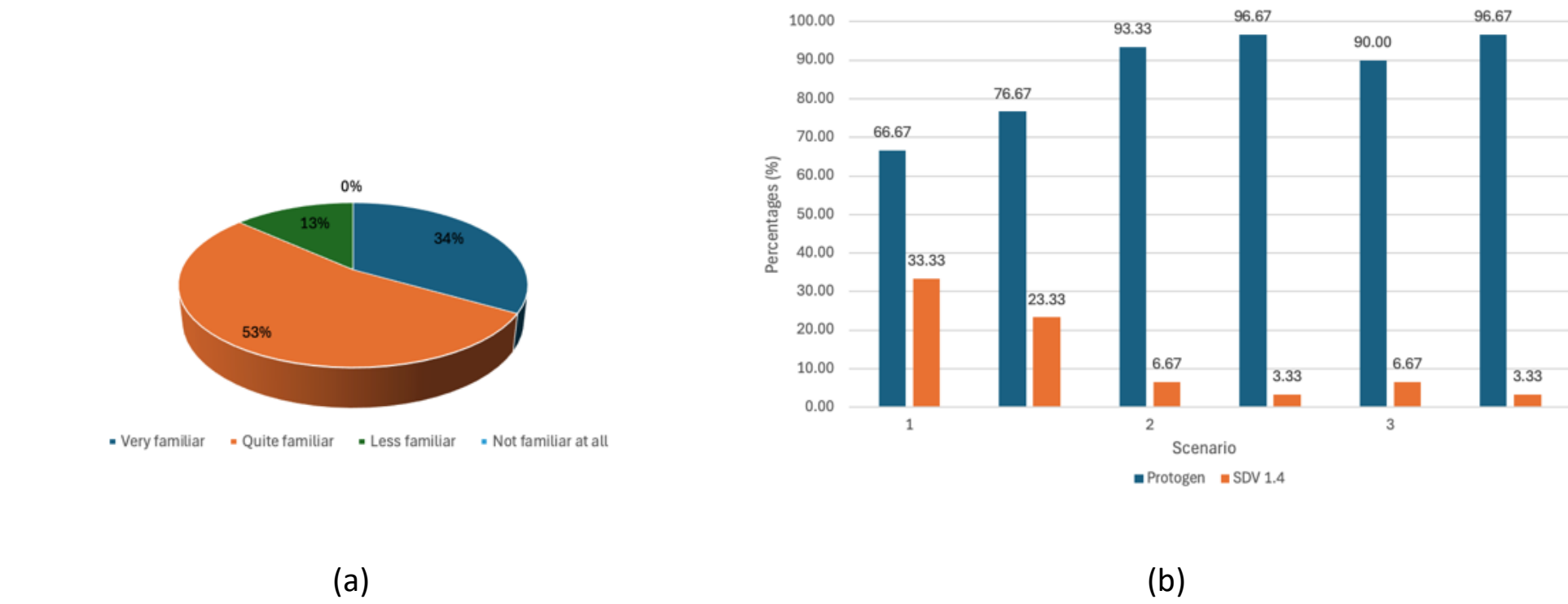

(a) (b)

Figure 8: Participants' level of familiarity with Ulos motifs (a), Quantitative responses from participants evaluating motifs generated by the Latent Diffusion Model (b)

Figure 8 shows that participant evaluations are conducted by respondents with strong familiarity with Ulos motifs; 86.6% report moderate to high familiarity, ensuring reliable assessments. The results consistently demonstrate a clear preference for motifs generated by Protogen x3.4 over SDV 1.4 across all three testing scenarios. Protogen is preferred by 66.67% of participants in Scenario 1, 93.33% in Scenario 2, and 96.67% in Scenario 3, indicating its superior ability to generate visually appealing, culturally representative, and aesthetically faithful Ulos motifs. These findings suggest that Protogen x3.4 captures structural patterns, symmetry, and cultural characteristics of traditional Ulos textiles more effectively, making it a more suitable model for heritage-inspired motif generation.

## 4 CONCLUSION

This study has demonstrated the capability of the Latent Diffusion Model (LDM) framework, enhanced through text-based conditioning, to generate novel digital interpretations of Batak Ulos motifs while preserving their core cultural and structural integrity. Across three experimental scenarios: structural modification, colour variation, and complex motif synthesis, the fine-tuned models successfully balanced visual novelty with cultural coherence, retaining traditional attributes such as bilateral symmetry, geometric precision, and the characteristic vertical layout intrinsic to Ulos textiles. These results confirm that diffusion-based generative frameworks can serve as effective instruments for both heritage preservation and design innovation, illustrating how traditional cultural expression can be meaningfully extended into the digital domain.

Quantitative evaluations across multiple configurations substantiate these findings. Across all tested scenarios, Protogen v3.4 consistently outperformed Stable Diffusion v1.4, achieving substantially lower Fréchet Inception Distance (FID) and Kernel Inception Distance (KID) values and higher Inception Scores (IS) throughout training. In Protogen v3.4 configurations, lower strength values consistently produced superior fidelity, as reflected in lower FID scores, while higher strength values enhanced generative diversity at the cost of reduced realism, revealing a clear fidelity–diversity trade-off rather than a single, universally optimal strength setting. Independent of strength, a guidance scale in the range of 5–9 provided the most effective balance between fidelity and diversity across all tested configurations, consistently stabilizing FID and IS for both models; this range is therefore recommended as the operating setting for high-quality, diverse Ulos motif generation. Complementary qualitative assessments involving traditional weavers and design students, conducted through structured interviews and questionnaires, further validated the practical relevance of the generated motifs at this configuration, confirming their weaving feasibility, aesthetic integrity, and cultural acceptability.

These findings position text-conditioned latent diffusion models as powerful computational tools for digitally preserving and revitalizing intangible cultural heritage. By bridging algorithmic precision with cultural symbolism, this approach demonstrates how artificial intelligence can help sustain the semantic depth of traditional designs while fostering contemporary creativity. Importantly, this work underscores AI's role as a collaborator rather than a replacement, a medium through which cultural craftsmanship can evolve within ethical, context-aware design frameworks that keep artisans and cultural custodians central to the creative process.

Future work will extend this foundation in several directions. The dataset will be broadened to encompass a wider range of Indonesian weaving traditions, capturing regional diversity in pattern structure, colour palette, and symbolic meaning. The generative framework will incorporate multimodal conditioning, integrating semantic maps, textual descriptions, historical narratives, weaving techniques, and material representations to achieve richer contextual awareness and finer control over motif synthesis. Human-in-the-loop evaluation systems involving local artisans, cultural scholars, and design practitioners will be developed to further refine cultural authenticity and support co-creative design. Finally, real-time co-creation interfaces could transform diffusion-based models into interactive design assistants, enabling artisans to collaborate dynamically with AI in crafting sustainable, culturally resonant textile innovations.

In conclusion, this research establishes a computational foundation for AI-assisted cultural preservation, demonstrating that latent diffusion models can mediate the intersection between tradition and technology. By aligning generative modelling with cultural understanding, this study contributes toward a paradigm of sustainable digital craftsmanship, in which human expertise and machine creativity converge to preserve, reinterpret, and advance the heritage of Indonesian textiles in the era of intelligent design.

## 5 AI USAGE DECLARATION

We acknowledge the use of Claude (Anthropic, Claude Sonnet4.6) and Grammarly to assist with proofreading, copy editing, and improving the clarity and flow of the manuscript. The authors maintain full responsibility for the content and final interpretation of the data.

## ACKNOWLEDGMENTS

This research was funded by the IT Del Research and Community Service Unit (LPPM), through the internal research funding program, Flagship Basic Research Scheme under Grant No 013.22/ITDel/LPPM/Penelitian/Ill/2025.